\documentclass[letterpaper]{article} 
\usepackage[preprint]{aaai2027}  
\usepackage[hyphens]{url}  
\usepackage{graphicx} 
\usepackage{amsmath}
\usepackage{natbib}  
\usepackage{caption} 
\usepackage{booktabs}

\title{QuISE: Defense against Typographic Attacks on VLMs \\ via Query-Irrelevant Semantic Editing}
\author{
Shubin Lu\textsuperscript{\rm 1},
Jiaqi Yin\textsuperscript{\rm 1},
Yihao Huang\textsuperscript{\rm 2}
}

\affiliations{
\textsuperscript{\rm 1}School of Software, Northwestern Polytechnical University, Xi'an, China\\
\textsuperscript{\rm 2}Software Engineering Institute, East China Normal University, Shanghai, China\\
lushubin@mail.nwpu.edu.cn, jqyin@nwpu.edu.cn, huangyihao@sei.ecnu.edu.cn
}

\begin{document}

\maketitle

\begin{abstract}
Typographic attacks pose a critical threat to vision-language models (VLMs) by injecting misleading text into images and causing models to rely on adversarial textual cues rather than visual evidence. Existing defenses often require model-specific modifications, additional training, or access to internal model components, limiting their applicability to modern closed-source VLMs. In this paper, we propose QuISE, a model-agnostic, training-free black-box defense based on query-irrelevant semantic editing. QuISE first identifies text regions likely to affect the current query through influence-aware text localization. QuISE then replaces these regions with two semantically distinct replacement texts that are irrelevant to both the query and the image. The final answer is determined by answer consistency across the edited images. Extensive experiments on three typographic-attack benchmarks, four attack settings, and four VLMs show that QuISE consistently improves defended accuracy. QuISE achieves a recovery rate of 67.9--75.0\% with a harm rate of 0.5--1.1\%.
\end{abstract}


\section{Introduction}
\label{sec:introduction}

Vision-language models (VLMs) use text embedded in images to answer queries about documents and charts~\cite{mathew2021docvqa,masry2022chartqa}, signs, and everyday scenes~\cite{singh2019textvqa,biten2019stvqa}. This capability also creates a semantic attack surface: typographic attacks insert misleading text that redirects answers away from visual evidence without changing the queried object. Unlike pixel-level perturbations, these attacks exploit readable semantics and remain effective in synthetic, real-world, and scene-coherent settings~\cite{westerhoff2025scam,cao2025scenetap}. A practical defense should therefore suppress the influence of misleading attack text without requiring access to the target VLM's internals.

\begin{figure}[t]
    \centering
    \includegraphics[width=\columnwidth]{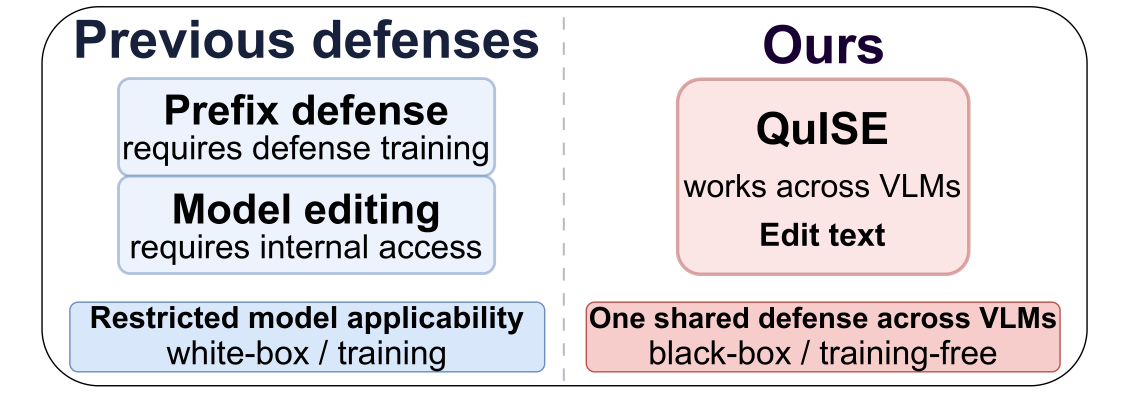}
    \caption{Comparison of existing defenses and QuISE.}
    \label{fig:defense_comparison}
\end{figure}

Existing defenses only partly meet this requirement. Defense-Prefix~\cite{azuma2023defenseprefix} and Dyslexify~\cite{hufe2026dyslexify} are specifically designed for CLIP-style models. Defense-Prefix learns model-specific prefixes that are tightly coupled with the CLIP representation space, while Dyslexify requires identifying and ablating internal attention heads in the CLIP vision encoder. Although effective for their target architectures, these approaches are difficult to generalize to closed-source VLMs, where model internals are inaccessible and the task paradigm extends beyond CLIP-style classification. Figure~\ref{fig:defense_comparison} contrasts existing defenses with QuISE's black-box design across VLMs.

To explore a black-box defense against typographic attacks on VLMs, we modify the inserted attack text to mitigate its effect. We first investigate which properties of attack text should be modified. Specifically, attack text has two fundamental aspects: appearance and semantics. Controlled modifications show that appearance modifications provide limited defense, whereas semantic modifications restore answer accuracy close to that on clean images.

We further investigate what semantics the attack text should be transformed into. Simply removing visible text is undesirable, because direct deletion may damage local visual evidence and disrupt the surrounding visual structure. Meanwhile, arbitrary semantic replacement may introduce new answer-related cues and remain influential to the model. Therefore, we analyze textual influence in the context of the complete multimodal input, where the effect of image text is determined by its interaction with both the query and the visual content. Specifically, we characterize replacement semantics from two perspectives: (1) query relevance, which measures whether the text directly provides information needed to answer the query; and (2) image relevance, which measures whether the text is supported by or consistent with the visual content. We compare replacements under the four semantic conditions formed by these two properties. The comparison shows that query relevance largely determines textual influence. Image relevance substantially affects the answer when the text is query-relevant, but has limited impact once the text becomes query-irrelevant. Consequently, query-irrelevant text is less likely to steer the model and is better suited for defensive replacement. Among the four conditions, Q0-I0, which is irrelevant to both the query and the image, provides the strongest defense.

Based on these observations, we propose \textbf{QuISE}, a model-agnostic, training-free black-box defense based on \emph{query-irrelevant semantic editing}. Without first determining whether an image is clean or attacked, QuISE applies two stages to each image--query pair. \emph{Influence-Aware Text Localization} combines target-VLM semantic judgment with auxiliary spatial grounding to identify text regions that may affect the current query. \emph{Query-Irrelevant Semantic Editing-based Defense} then creates two edited images using semantically distinct replacements verified to be irrelevant to both the query and the image. QuISE adopts an edited answer only when both edited images yield the same valid answer. Otherwise, it uses the target VLM's answer on the input image. This consensus-based selection supports answer recovery under attack.

The main contributions of this paper are as follows:
\begin{itemize}
    \item To the best of our knowledge, we present the first black-box defense against typographic attacks on VLMs via semantic editing. Our design stems from two key observations: semantic modifications outperform appearance modifications against the evaluated attacks, and query relevance largely determines textual influence.
    \item We propose QuISE, a model-agnostic, training-free black-box defense combining influence-aware localization, query-irrelevant semantic editing and consensus-based answer selection.
    \item We evaluate QuISE across three attack benchmarks and four open-weight and closed-source VLMs, and show that it outperforms three representative defenses.
\end{itemize}

\section{Related Work}
\label{sec:related_work}

\subsection{Visual Text Understanding in VLMs}

VLMs have progressed from contrastive image--text representation learning to general-purpose visual assistants.
LLaVA~\cite{liu2023llava} introduced visual instruction tuning for open-ended question answering and multimodal dialogue. Subsequent models have further strengthened fine-grained perception and text-rich image understanding. The Qwen-VL family~\cite{bai2023qwenvl,bai2025qwen25vl} supported visual grounding, text reading, high-resolution perception, and document parsing, while LLaVA-OneVision~\cite{li2025llavaonevision} and InternVL~\cite{chen2024internvl} extended multimodal reasoning across diverse visual tasks and input settings. Within this broader progression, interpreting image text has become an important component of VLM reasoning. TextVQA~\cite{singh2019textvqa} and ST-VQA~\cite{biten2019stvqa} evaluated whether models could answer queries by jointly reasoning over scene text and visual context, whereas OCRBench~\cite{liu2024ocrbench} broadened this evaluation to text recognition, scene-text VQA, and document understanding. 

\subsection{Typographic Attacks on VLMs}

Typographic attacks exploit the tendency of VLMs to treat image text as semantic evidence. Multimodal Neurons~\cite{goh2021multimodal} demonstrated that misleading labels could override depicted content in CLIP classification. Disentangling Visual and Written Concepts in CLIP~\cite{materzynska2022disentangling} related this behavior to the entanglement of written and visual concepts in the image encoder. Recent studies have extended the threat to generative VLMs and more realistic settings. Self-generated Typographic Attacks~\cite{qraitem2024selfgenerated} used an LVLM to produce context-dependent misleading text, while SCAM~\cite{westerhoff2025scam} evaluated both synthetic and physically captured attacks. SceneTAP~\cite{cao2025scenetap} jointly planned adversarial text and its placement to produce scene-coherent attacks. Typographic Attacks in a Multi-Image Setting~\cite{wang2025multiimage} improved stealth by selecting diverse, non-repeating attack words across an image set. Image text can also redirect model behavior as an instruction. FigStep~\cite{gong2025figstep} converted harmful instructions into typographic visual prompts to bypass text-side safety alignment, whereas Goal Hijacking via Visual Prompt Injection~\cite{kimura2024goalhijacking} embedded an alternative task in the image to replace the model's objective. These studies show that typographic attacks affect not only classification and query answering, but also instruction following and multimodal safety.

\subsection{Defense against Typographic Attacks}

Only a limited number of defense approaches have been proposed for typographic attacks. Defense-Prefix~\cite{azuma2023defenseprefix} learned a special prefix token placed before class names, making CLIP-based classifiers less sensitive to misleading words in the image. PAINT~\cite{ilharco2022paint} fine-tuned an open-vocabulary model on a target patching task and interpolated the original and fine-tuned weights to improve the target capability while preserving general performance. Dyslexify~\cite{hufe2026dyslexify} identified attention heads that causally transmitted typographic information through the CLIP vision encoder and selectively ablated them without retraining. Defense-Prefix depends on a CLIP-style class-name interface, whereas PAINT and Dyslexify require access to model weights or internal components. These requirements limit their direct application to open-ended and closed-source VLMs. QuISE instead localizes query-conditioned influential text, applies semantically distinct replacements verified to be irrelevant to both the query and the image, and changes the input-image answer only when the two edited answers form a valid consensus. This design enables a training-free input-level defense without access to the target model's internal components.

\section{Motivation}
\label{sec:motivation}
To inform the design of a black-box defense against typographic attacks, we examine how modifications to inserted attack text affect VLM answers. We first compare appearance modifications with semantic modifications, and then identify which replacement semantics most effectively weaken the influence of attack text.

\subsection{Change What: Appearance or Semantics?}
\label{sec:motivation_text_mutation}

\paragraph{Setup.}
We sample 600 attacked image--query pairs, with 200 each from SCAM, SceneTAP, and SELF. Each modification is applied only within the inserted-text region. Qwen2.5-VL-7B-Instruct and LLaVA-OneVision-8B answer the same query on each resulting image.

\begin{figure}[t]
    \centering
    \includegraphics[width=0.95\columnwidth]{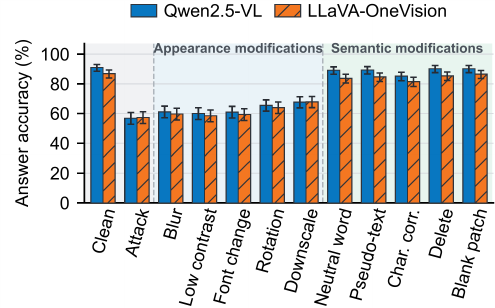}
    \caption{Acc under appearance \& semantic modifications.}
    \label{fig:text_mutation_motivation}
\end{figure}

\emph{Clean} denotes the image before attack text is added, whereas \emph{Attack} denotes the corresponding image containing the misleading attack text. The five appearance modifications preserve the semantics of the attack text: \emph{Blur}, \emph{Low contrast}, \emph{Font change}, \emph{Rotation}, and \emph{Downscale}. The five semantic modifications alter or remove the textual meaning: \emph{Neutral word} replaces attack text with a more abstract concept (e.g., ``pen'' $\rightarrow$ ``object''); \emph{Pseudo-text} replaces it with meaningless characters; \emph{Character corruption} randomly replaces characters with symbols; \emph{Delete} removes the text region; and \emph{Blank patch} fills the region with background color.

\textbf{Observation 1: Semantic modifications provide a stronger defense than appearance modifications.}
Figure~\ref{fig:text_mutation_motivation} compares the two modification families across both VLMs. Attacked images achieve an average accuracy of 57.08\%. Appearance modifications improve the accuracy to 64.64\%. Semantic modifications substantially raise it to 86.57\%, approaching the 88.92\% accuracy on clean images. This indicates that the semantic content of inserted text is the primary factor driving typographic attacks.

\subsection{Which Semantics for Defense?}
\label{sec:motivation_semantic_intervention}

\begin{figure}[t]
    \centering
    \includegraphics[width=\columnwidth]{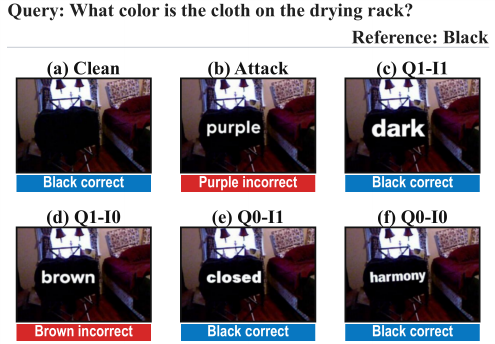}
    \caption{Example of the four controlled semantic modifications and the corresponding VLM answers.}
    \label{fig:semantic_intervention_example}
\end{figure}

\paragraph{Semantic properties.}
Observation 1 shows that semantic modification is more effective than appearance modification. We therefore study which replacement semantics can minimize the influence of inserted text. Specifically, we characterize semantics along two dimensions: query relevance (\textbf{Q}) and image relevance (\textbf{I}), where 1 and 0 denote relevance and irrelevance, yielding four conditions: \textbf{Q1-I1} (relevant to both), \textbf{Q1-I0} (query-relevant only), \textbf{Q0-I1} (image-relevant only), and \textbf{Q0-I0} (irrelevant to both). Figure~\ref{fig:semantic_intervention_example} illustrates these conditions with an example.

\paragraph{Setup.}
We generate all four verified semantic variants for 1,810 image--query pairs: 1,010 from SCAM and 800 from SceneTAP. For each pair, we evaluate the \emph{Clean} and \emph{Attack} versions, a version with the ground-truth answer inserted as image text, and the four semantic variants. The \emph{Clean} and \emph{Attack} versions serve as the pre-attack and attacked references, respectively. The four semantic variants are evaluated as controlled defense modifications.

\begin{figure}[t]
    \centering
    \includegraphics[width=\columnwidth]{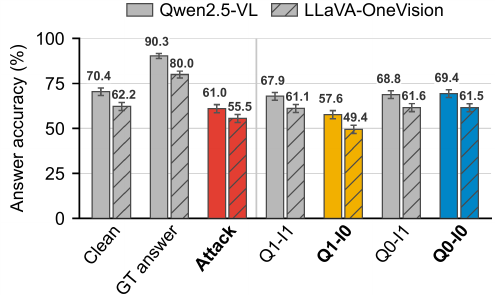}
    \caption{Acc under four semantic conditions.}
    \label{fig:semantic_intervention_results}
\end{figure}
\begin{figure*}[t]
    \centering
    \includegraphics[width=0.9\textwidth]{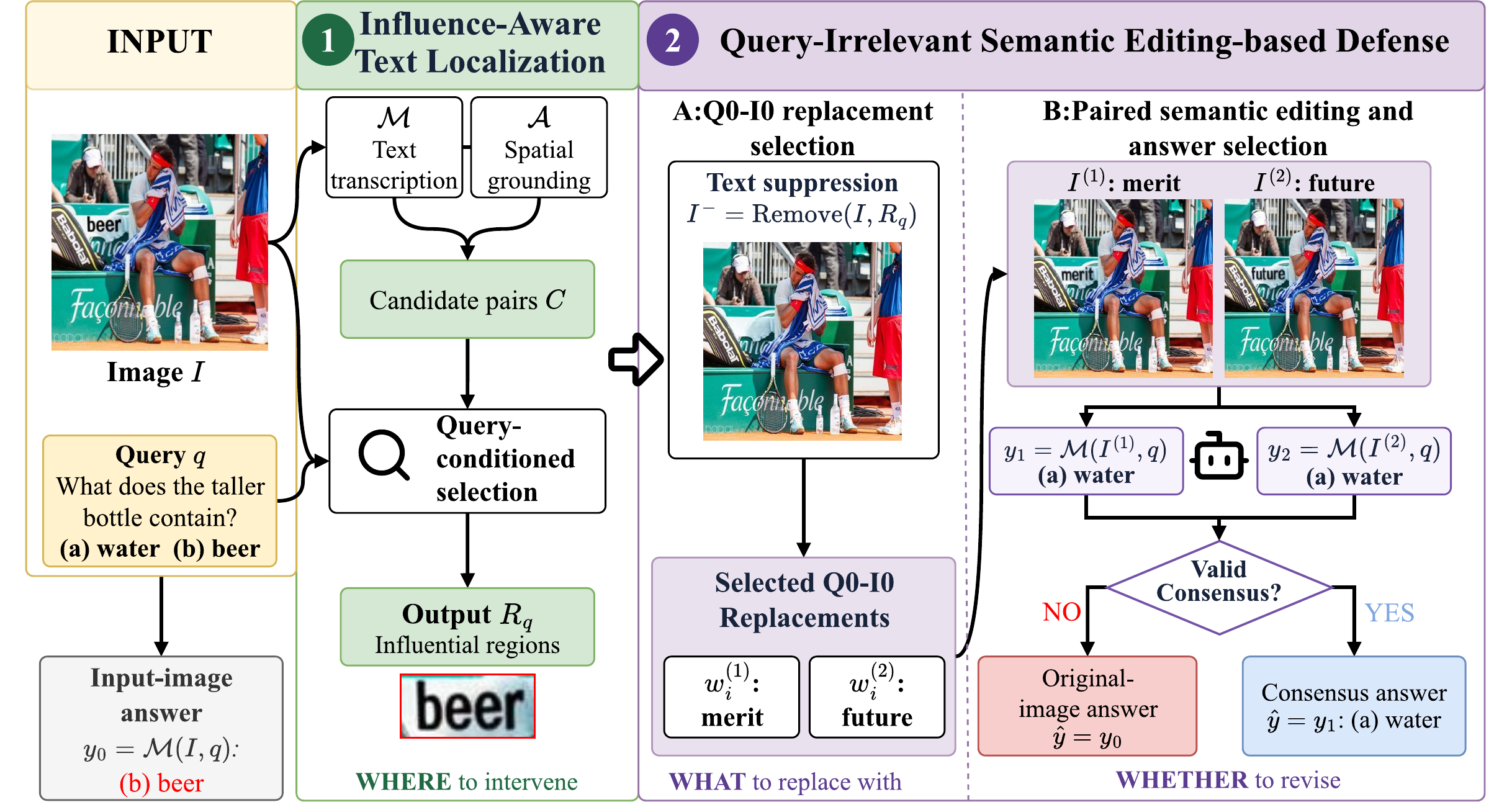}
    \caption{Overview of QuISE.}
    \label{fig:framework}
\end{figure*}

\paragraph{Observation 2: Query relevance largely determines textual influence.}
Figure~\ref{fig:semantic_intervention_results} compares the four semantic conditions. For query-relevant replacements, Q1-I1 consistently outperforms Q1-I0 by 10.28 and 11.66 points on Qwen2.5-VL and LLaVA-OneVision, respectively. In contrast, Q0-I1 and Q0-I0 achieve comparable performance, indicating that image relevance has limited impact once the text is query-irrelevant. These results show that query relevance is the primary factor determining whether inserted text influences VLM decisions. Among the four conditions, Q0-I0 provides the strongest defense. It improves accuracy over the \emph{Attack} condition by 8.40 and 6.02 points on Qwen2.5-VL and LLaVA-OneVision, respectively, while approaching \emph{Clean} performance. We therefore select Q0-I0 as the target replacement semantics for QuISE.

\section{Method}
\label{sec:method}

\subsection{Problem Formulation}
\label{sec:problem_formulation}

Given an image--query pair $(I,q)$, let $\mathcal{M}$ denote the target VLM, which produces:
\begin{equation}
    y_0=\mathcal{M}(I,q).
\end{equation}
The input image may contain visible text. It may also contain attack text inserted by a typographic attack. We call $I$ a clean image if it contains no attack text, and an attacked image if it contains attack text. At inference, the defense only observes $(I,q)$ and does not know whether $I$ is clean or attacked.

We aim to design a defense $\mathcal{D}$ that outputs a correct answer:
\begin{equation}
    \hat{y}=\mathcal{D}(\mathcal{M};I,q),
\end{equation}
where the ground-truth answer $y^*$ is evaluation-only. An effective defense should suppress attack-induced textual influence without indiscriminately editing all visible text regions.

We consider a black-box setting where $\mathcal{D}$ can only interact with $\mathcal{M}$ through input manipulation and output analysis, without access to model parameters, gradients, probabilities, or intermediate representations.

\subsection{Overview}
\label{sec:overview}
Figure~\ref{fig:framework} shows the two stages of QuISE. First, \emph{Influence-Aware Text Localization} identifies text regions that may affect the answer to the current query. Second, \emph{Query-Irrelevant Semantic Editing-based Defense} replaces each localized region with two verified Q0-I0 replacements to generate two edited images. QuISE accepts an edited answer only if both edited images yield consistent responses.

\subsection{Influence-Aware Text Localization}
\label{sec:influence_localization}

Not every visible text region affects the answer to the current query, so it is unnecessary to edit all detected text. QuISE identifies text regions that may influence the target model's answer and edits only the selected regions.
Specifically, QuISE first extracts candidate text regions from the image. The target model $\mathcal{M}$ is used to recognize visible text, while a fixed auxiliary component $\mathcal{A}$ provides the corresponding spatial locations. The resulting candidate set is represented as:
\begin{equation}
    C=\operatorname{Associate}_{\mathcal{M},\mathcal{A}}(I)
    =\{(l_j,t_j)\}_{j=1}^{N},
    \label{eq:candidate_regions}
\end{equation}
where $C$ contains $N$ candidate location--text pairs. For the $j$-th candidate, $l_j$ denotes the text location and $t_j$ denotes its content. The candidate set only describes visible text and does not indicate whether a region affects the current query. Given $(I,q,C)$, QuISE uses the target model to select influential regions:
\begin{equation}
    R_q=\operatorname{Select}_{\mathcal{M}}(I,q,C)
    =\{(l_i,t_i)\}_{i=1}^{K}\subseteq C,
    \label{eq:localized_regions}
\end{equation}
where $R_q$ contains $K$ selected location--text pairs for subsequent semantic editing.

\subsection{Query-Irrelevant Semantic Editing-based Defense}

\paragraph{Q0-I0 replacement selection.}
After locating potentially influential text regions, QuISE edits their semantics. Direct removal may discard useful visual information and disrupt the local structure. Arbitrary replacement may introduce new answer-related cues. QuISE therefore replaces the text in each localized region with content irrelevant to both the current query and the image. This operation reduces the influence of potentially misleading semantics while preserving the visual form and layout of the selected regions as much as possible.

To obtain suitable replacements, QuISE first removes the localized text regions $R_q$:
\begin{equation}
    I^{-}=\mathrm{Remove}(I,R_q),
    \label{eq:text_removed_image}
\end{equation}
$I^{-}$ is the text-removed image for evaluating image relevance.

For each localized region, QuISE uses the target model $\mathcal{M}$ to generate candidate replacements conditioned on the query $q$ and the text-removed image $I^{-}$. It retains only candidates that are irrelevant to both the query and the image, referred to as Q0-I0 replacements. For the $i$-th localized region, QuISE selects two replacements with different semantics:
\begin{equation}
    P_i=(w_i^{(1)},w_i^{(2)}),
    \qquad i=1,\ldots,K,
    \label{eq:replacement_pairs}
\end{equation}
where $K$ is the number of localized regions, and $w_i^{(1)}$ and $w_i^{(2)}$ denote the two replacements selected for region $i$.

\paragraph{Paired semantic editing and answer selection.}

Using the replacement pairs, QuISE generates two edited images:
\begin{equation}
    (I^{(1)},I^{(2)})
    =\mathrm{Edit}\!\left(I,R_q,\{P_i\}_{i=1}^{K}\right),
    \label{eq:edited_images}
\end{equation}
where each edited image applies one replacement from every pair. The target model produces:
\begin{equation}
    y_1=\mathcal{M}(I^{(1)},q),\qquad
    y_2=\mathcal{M}(I^{(2)},q).
    \label{eq:edited_answers}
\end{equation}

QuISE accepts the edited answer only when both answers are valid and consistent:
\begin{equation}
    \hat{y}=
    \begin{cases}
        y_1, & y_1,y_2\text{ are valid and }y_1=y_2,\\
        y_0, & \text{otherwise},
    \end{cases}
    \label{eq:paired_decision}
\end{equation}
where $y_0=\mathcal{M}(I,q)$ is the target model's answer on the input image. An answer is valid if it can be parsed into the task-specific answer space and is neither empty nor a refusal.

\paragraph{Preserving clean-image predictions.}
QuISE is designed to limit unnecessary answer changes on clean images. If no candidate text region is identified or no region is selected, QuISE skips editing and retains the input-image answer $y_0$. On a clean image, a selected region can contain text whose content is needed to answer the query, such as a sign label or a chart value. The two edited images replace this text with different Q0-I0 replacements. If the answer depends on the selected text, the edited answers may differ or become invalid. No valid consensus is then formed, and QuISE retains $y_0$. This fallback can limit performance loss on clean images, including queries that require reading visible text.

\begin{table*}[t]
\centering
\small
\setlength{\tabcolsep}{0.9mm}
\renewcommand{\arraystretch}{1.08}
\begin{tabular}{@{}ll|rrrrr|rr@{}}
\toprule
Target model & Defense method & SCAM-S & SCAM-R & SceneTAP & SELF & Overall & RR$\uparrow$ & HR$\downarrow$ \\
\midrule
OpenAI CLIP ViT-B/32 & Defense-Prefix & 63.1 (+20.2) & 79.3 (+13.0) & 58.8 (+14.8) & 47.0 (+14.7) & 55.7 (+15.2) & 39.3 & 3.2 \\
\addlinespace[2pt]
OpenCLIP ViT-B/16 & Dyslexify & 83.8 (+30.7) & 89.0 (+18.8) & 55.2 (+14.7) & 57.2 (+33.3) & 64.8 (+27.6) & 57.9 & 1.9 \\
\midrule
Qwen2.5-VL-7B-Instruct & Attacked & 89.2 & 89.7 & 71.9 & 17.1 & 46.8 & -- & -- \\
& AAP & 97.5 (+8.3) & 96.6 (+6.9) & 83.3 (+11.4) & 19.0 (+1.9) & 52.0 (+5.2) & 18.8 & 6.3 \\
& \textbf{QuISE} & \textbf{99.2 (+10.0)} & \textbf{98.6 (+9.0)} & \textbf{89.5 (+17.6)} & \textbf{63.7 (+46.5)} & \textbf{78.0 (+31.2)} & \textbf{73.9} & \textbf{0.8} \\
\addlinespace[2pt]
LLaVA-OneVision-8B & Attacked & 86.8 & 92.9 & 72.3 & 11.3 & 43.8 & -- & -- \\
& AAP & 88.3 (+1.5) & 86.9 ($-$6.0) & 79.6 (+7.4) & 14.4 (+3.1) & 46.2 (+2.4) & 19.9 & 13.0 \\
& \textbf{QuISE} & \textbf{98.6 (+11.8)} & \textbf{98.7 (+5.8)} & \textbf{92.7 (+20.5)} & \textbf{54.2 (+42.9)} & \textbf{73.3 (+29.5)} & \textbf{75.0} & \textbf{0.5} \\
\addlinespace[2pt]
InternVL3.5-8B & Attacked & 67.6 & 78.7 & 59.3 & 4.5 & 33.2 & -- & -- \\
& AAP & 91.0 (+23.4) & 89.4 (+10.8) & 80.7 (+21.4) & 10.7 (+6.2) & 45.2 (+11.9) & 29.2 & 2.3 \\
& \textbf{QuISE} & \textbf{96.0 (+28.5)} & \textbf{95.6 (+17.0)} & \textbf{87.3 (+28.1)} & \textbf{39.2 (+34.7)} & \textbf{63.4 (+30.2)} & \textbf{70.3} & \textbf{1.1} \\
\addlinespace[2pt]
GPT-4.1-mini & Attacked & 99.0 & 99.1 & 86.8 & 54.8 & 72.7 & -- & -- \\
& AAP & \textbf{99.7 (+0.8)} & \textbf{99.7 (+0.5)} & \textbf{93.4 (+6.6)} & 50.8 ($-$4.0) & 71.9 ($-$0.8) & 18.8 & 7.0 \\
& \textbf{QuISE} & 99.4 (+0.4) & \textbf{99.7 (+0.5)} & 92.4 (+5.6) & \textbf{76.7 (+21.9)} & \textbf{85.8 (+13.1)} & \textbf{67.9} & \textbf{0.8} \\
\bottomrule
\end{tabular}
\caption{Defense effectiveness under typographic attacks (\%). SCAM-S/R denote synthetic/real-world SCAM conditions; parentheses show gains over attacked inputs. Bold marks the better of AAP and QuISE within each target model. Defense-Prefix and Dyslexify use their native CLIP backbones.}
\label{tab:attack_overall}
\end{table*}
\begin{figure*}[t]
    \centering
    \includegraphics[width=0.8\textwidth]
    {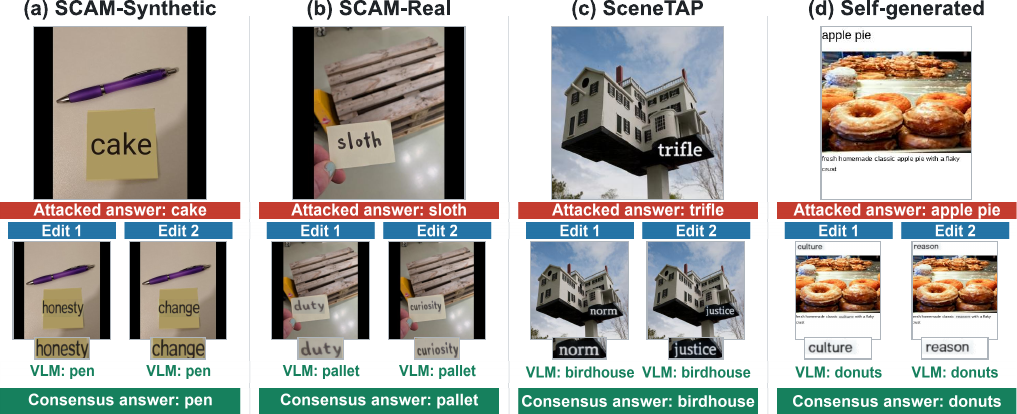}
    \caption{Qualitative examples of QuISE. For each attacked image, two query-irrelevant edits lead to a consistent recovered answer. Red and green bars denote the attacked and recovered consensus answers, respectively.}
    \label{fig:quise-qualitative}
\end{figure*}

\section{Experiments}
\label{sec:experiments}
We primarily evaluate QuISE's defense against typographic attacks. We also assess whether it preserves performance on clean images. A clean image contains no attack text. It may contain no visible text or text unrelated to the query. It may also contain text that must be read to answer the query.

\subsection{Experimental Settings}
\label{sec:experimental_settings}

\paragraph{Datasets.}
We evaluate defense effectiveness on SCAM~\cite{westerhoff2025scam}, SceneTAP~\cite{cao2025scenetap}, and the self-generated typographic attack benchmark (SELF)~\cite{qraitem2024selfgenerated}. SCAM contains 1,162 image--query pairs with both synthetic and real-world attacked variants, which we report separately. SceneTAP and SELF contain 1,572 and 4,662 attacked pairs, respectively. Each target model is therefore evaluated on 8,558 attacked instances and their clean counterparts. To evaluate performance on clean text-reading tasks, we use fixed subsets of 500 unique image--query pairs from each of TextVQA~\cite{singh2019textvqa} and ST-VQA~\cite{biten2019stvqa}. These images contain no attack text, and each query requires reading visible text to obtain the answer.

\paragraph{Target models.}
We evaluate our method on Qwen2.5-VL-7B-Instruct~\cite{bai2025qwen25vl}, LLaVA-OneVision-8B~\cite{li2025llavaonevision}, InternVL3.5-8B~\cite{wang2025internvl35}, and GPT-4.1-mini~\cite{openai2025gpt41}. 

\paragraph{Baselines.}
We compare QuISE with representative defenses. Artifact-aware Prompting (AAP)~\cite{qraitem2025webartifact} is a prompt-based mitigation that first asks the VLM to inspect suspicious text or logos and then answers the query using the inspection context. Defense-Prefix~\cite{azuma2023defenseprefix} learns defensive prefixes to reduce CLIP's sensitivity to typographic text, while Dyslexify~\cite{hufe2026dyslexify} removes attention heads responsible for textual information transmission in CLIP vision encoders. We further include Localized Deletion, which shares QuISE's localization stage but removes text regions instead of editing their semantics, to evaluate the contribution of semantic editing. 

AAP is evaluated on the same target models and instances as QuISE. Since Defense-Prefix and Dyslexify rely on CLIP-family backbones, we compare their defense gains, RR, and HR on the same benchmark inputs while noting the different target models. CLIP-based defenses are excluded from TextVQA and ST-VQA because they do not support open-ended scene-text question answering.

\paragraph{Metrics.}
Following prior typographic-attack evaluations~\cite{westerhoff2025scam,cao2025scenetap,qraitem2024selfgenerated}, we use answer accuracy as the primary metric. We report attacked accuracy $\mathrm{Acc}_{\mathrm{att}}$, defended accuracy $\mathrm{Acc}_{\mathrm{def}}$, and the improvement $\Delta=\mathrm{Acc}_{\mathrm{def}}-\mathrm{Acc}_{\mathrm{att}}$.

To distinguish attack recovery from defense-induced degradation, we additionally report Recovery Rate (RR) and Harm Rate (HR). For sample $i$, let $c_i,a_i,d_i\in\{0,1\}$ denote the correctness of the clean, attacked, and defended answers, respectively. RR measures the fraction of attack-induced errors corrected by the defense:
\begin{equation}
    \mathrm{RR}
    =100\,
    \frac{
        |\{i:c_i=1,a_i=0,d_i=1\}|
    }{
        |\{i:c_i=1,a_i=0\}|
    }.
\end{equation}
HR measures the fraction of originally correct attacked answers that become incorrect after defense:
\begin{equation}
    \mathrm{HR}
    =100\,
    \frac{
        |\{i:a_i=1,d_i=0\}|
    }{
        |\{i:a_i=1\}|
    }.
\end{equation}
Higher accuracy and RR, together with lower HR, indicate better defense performance. We report overall results across 8,558 attacked instances and per-setting results.

For these clean text-reading tasks, we evaluate TextVQA using VQA soft accuracy and ST-VQA using the average normalized Levenshtein similarity (ANLS) with a threshold of 0.5. For each defense policy
$r$, we compute the macro-average score $M_r=(S_r^{\mathrm{TextVQA}}+S_r^{\mathrm{ST\text{-}VQA}})/2$
and define performance retention as $\mathrm{Ret}_r=100M_r/M_0$, where $M_0$ is the corresponding macro-average score without defense.

\paragraph{Implementation.}
We use PP-OCRv6 in PaddleOCR 3.7.0~\cite{zhang2026ppocrv6} as the auxiliary text-localization component, LaMa~\cite{suvorov2022lama} to remove the localized text, and RS-STE~\cite{fang2025rsste} to render replacement text. We use the same components, prompts, and hyperparameters across all target models and attack datasets. We run all local inference on a single NVIDIA vGPU-32GB with 32\,GB of memory and access GPT-4.1-mini through its API. If any defense stage fails, we retain the input-image answer and include the sample in all metrics. Further implementation details are provided in the supplementary material.

\subsection{Qualitative Analysis}
Figure~\ref{fig:quise-qualitative} presents representative defense cases across four attack settings. Query-irrelevant edits suppress the misleading text and produce the same correct answer.

\subsection{Defense Effectiveness}
\label{sec:attack_results}

Table~\ref{tab:attack_overall} compares QuISE with representative defenses to evaluate its effectiveness against typographic attacks. For each target model, we report the attacked accuracy, defended accuracy, and the improvement over attacked inputs in parentheses. We additionally report RR and HR, which measure the ability to recover attack-induced errors and the risk of introducing new errors, respectively. Higher accuracy and RR, together with lower HR, indicate better defense performance.

QuISE consistently outperforms existing defenses overall across target models and attack settings. Compared with AAP, which is the strongest VLM-based baseline, QuISE improves overall defended accuracy by 13.9--27.1 percentage points, achieving 67.9--75.0\% RR with only 0.5--1.1\% HR. In contrast, AAP achieves only 18.8--29.2\% RR and introduces higher HRs of 2.3--13.0\%. These results show that semantic editing is substantially more effective than prompt-based mitigation for removing misleading textual influence.

QuISE also outperforms CLIP-based defenses. Defense-Prefix and Dyslexify achieve 39.3\% and 57.9\% RR with 3.2\% and 1.9\% HR, respectively, whereas QuISE achieves higher RR and lower HR. Although these methods rely on different CLIP-family backbones, the comparison on the same benchmark inputs shows that QuISE provides stronger attack recovery without requiring model-specific modification.

The advantage of QuISE is particularly evident under stronger attacks. On SELF, where attacked accuracy is substantially lower, QuISE improves accuracy by 21.9--46.5 points, while AAP provides limited gains and even decreases GPT-4.1-mini accuracy. These results demonstrate that directly modifying misleading text semantics is more effective than relying only on prompting when VLMs are strongly affected by typographic attacks.

\begin{figure}[t]
    \centering
    \includegraphics[width=0.8\columnwidth]{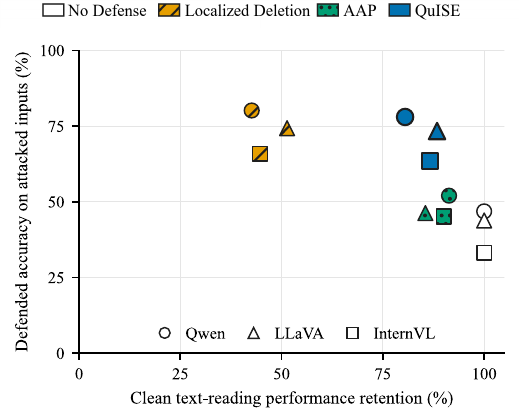}
    \caption{Defense effectiveness and performance retention on clean text-reading tasks for three open-weight VLMs.}
    \label{fig:defense_utility}
\end{figure}

\subsection{Effectiveness on Clean Text-Reading Tasks}
\label{sec:text_utility}

Figure~\ref{fig:defense_utility} compares QuISE with AAP and Localized Deletion. The comparison considers defense effectiveness against typographic attacks and performance retention on clean text-reading tasks. Localized Deletion achieves strong defense performance by removing selected text regions but may also remove text needed to answer the query. QuISE remains within 0.9--2.5 accuracy points of Localized Deletion while achieving 37.0--42.0 percentage points higher performance retention. Compared with AAP, QuISE improves defense effectiveness on all target models. These results show that QuISE better balances defense effectiveness with performance on clean text-reading tasks.

\begin{table}[t]
\centering
{\small
\setlength{\tabcolsep}{1mm}
\renewcommand{\arraystretch}{1.08}
\begin{tabular}{@{}lrrrrr@{}}
\toprule
Policy & Acc.$\uparrow$ & RR$\uparrow$ & HR$\downarrow$
& Ret.$\uparrow$ & Edits$\downarrow$ \\
\midrule
No Defense
& 46.83 & -- & -- & 100.00 & 0 \\

Single Edit (avg.)
& 79.38 & 76.48 & 0.94 & 40.46 & 1 \\

Three-Edit Majority
& \textbf{79.47} & \textbf{76.70} & 0.92 & 76.21 & 3 \\

Three-Edit Unanimous
& 77.37 & 72.43 & \textbf{0.72} & \textbf{83.24} & 3 \\

\textbf{QuISE}
& 78.00 & 73.85 & 0.80 & 80.49 & 2 \\
\bottomrule
\end{tabular}
}
\caption{Edit-count and consensus analysis.}
\label{tab:edit_count_consensus}
\end{table}

\subsection{Analysis of Edit Count and Consensus Policy}
\label{sec:edit_count_ablation}

We analyze how the number of edited images and consensus policy affect QuISE on Qwen2.5-VL-7B-Instruct, with all other components fixed. For Single Edit, we separately evaluate three fixed one-edit settings and average their dataset-level metrics. Three-Edit Majority accepts an answer when at least two edited answers agree, whereas Three-Edit Unanimous requires all three answers to agree. QuISE uses two edited images and accepts the edited answer only when both answers agree. Otherwise, it retains the input-image answer.

As shown in Table~\ref{tab:edit_count_consensus}, Single Edit achieves 76.48\% RR. However, it achieves only 40.46\% performance retention on the clean text-reading tasks. This result indicates that consensus verification substantially improves performance retention. Compared with QuISE, Three-Edit Majority slightly improves accuracy and RR but requires an additional edited image and target-model query. Three-Edit Unanimous increases performance retention to 83.24\% and reduces HR to 0.72\%, but lowers RR to 72.43\%. QuISE achieves 73.85\% RR, 0.80\% HR, and 80.49\% performance retention using only two edited images. These results show that two-edit consensus balances attack recovery, performance retention on clean text-reading tasks, and inference cost.

\subsection{Performance on Clean Recognition Tasks Not Requiring Image Text}
\label{sec:clean_recognition}

We further evaluate QuISE on clean recognition tasks not requiring image text. We use fixed multiple-choice subsets of FGVC-Aircraft~\cite{maji2013aircraft}, Food-101~\cite{bossard2014food101}, and ImageNet-100~\cite{shekhar2021imagenet100}, with 1,001, 1,001, and 1,000 samples.

As shown in Table~\ref{tab:clean_recognition}, QuISE causes no accuracy drop on any model--dataset pair and introduces no harmful correct-to-incorrect flips across 9,006 predictions. The paired-consensus and fallback rule preserves target-model predictions on the evaluated clean-image recognition tasks.

\begin{table}[h]
\centering
{\small
\setlength{\tabcolsep}{1mm}
\renewcommand{\arraystretch}{1.08}
\begin{tabular}{@{}lrrr@{}}
\toprule
& \multicolumn{3}{c}{Clean Images} \\
\cmidrule(lr){2-4}
Model & Aircraft & Food-101 & ImageNet-100 \\
\midrule
Qwen2.5-VL
& 64.64 (0.00) & 75.72 (0.00) & 98.70 (0.00) \\

LLaVA-OneVision
& 58.94 (0.00) & 77.32 (+0.10) & 98.80 (0.00) \\

InternVL3.5
& 41.46 (0.00) & 67.63 (0.00) & 98.40 (0.00) \\
\bottomrule
\end{tabular}
}
\caption{Accuracy on clean recognition tasks not requiring image text. Parentheses show changes from No Defense.}
\label{tab:clean_recognition}
\end{table}

\section{Conclusion}
We investigated defense against typographic attacks on vision-language models and showed that attack effectiveness mainly stems from the semantic content of inserted text rather than its visual appearance. Based on this insight, we proposed QuISE, a model-agnostic, training-free black-box defense that mitigates misleading textual influence. Extensive experiments demonstrate that QuISE achieves strong attack recovery across diverse VLMs and attack settings while preserving clean-image performance.

\bibliography{aaai2027}

@inproceedings{liu2023llava,
  title     = {Visual Instruction Tuning},
  author    = {Liu, Haotian and Li, Chunyuan and Wu, Qingyang and Lee, Yong Jae},
  booktitle = {Advances in Neural Information Processing Systems},
  volume    = {36},
  pages     = {34892--34916},
  year      = {2023},
  doi       = {10.52202/075280-1516}
}

@misc{bai2023qwenvl,
  title         = {{Qwen-VL}: A Versatile Vision-Language Model for Understanding, Localization, Text Reading, and Beyond},
  author        = {Bai, Jinze and Bai, Shuai and Yang, Shusheng and Wang, Shijie and Tan, Sinan and Wang, Peng and Lin, Junyang and Zhou, Chang and Zhou, Jingren},
  year          = {2023},
  eprint        = {2308.12966},
  archivePrefix = {arXiv},
  primaryClass  = {cs.CV},
  doi           = {10.48550/arXiv.2308.12966}
}

@article{westerhoff2025scam,
  title   = {{SCAM}: A Real-World Typographic Robustness Evaluation for Multimodal Foundation Models},
  author  = {Westerhoff, Justus and Purelku, Erblina and Hackstein, Jakob and Loos, Jonas and Pinetzki, Leo and Rodner, Erik and Hufe, Lorenz},
  journal = {Journal of Data-centric Machine Learning Research},
  volume  = {3},
  number  = {4},
  pages   = {1--30},
  year    = {2026},
  url     = {https://openreview.net/forum?id=zCcJSErVHH}
}

@inproceedings{cao2025scenetap,
  title     = {{SceneTAP}: Scene-Coherent Typographic Adversarial Planner against Vision-Language Models in Real-World Environments},
  author    = {Cao, Yue and Xing, Yun and Zhang, Jie and Lin, Di and Zhang, Tianwei and Tsang, Ivor and Liu, Yang and Guo, Qing},
  booktitle = {Proceedings of the IEEE/CVF Conference on Computer Vision and Pattern Recognition},
  pages     = {25050--25059},
  year      = {2025}
}

@inproceedings{singh2019textvqa,
  title     = {Towards {VQA} Models That Can Read},
  author    = {Singh, Amanpreet and Natarajan, Vivek and Shah, Meet and Jiang, Yu and Chen, Xinlei and Batra, Dhruv and Parikh, Devi and Rohrbach, Marcus},
  booktitle = {Proceedings of the IEEE/CVF Conference on Computer Vision and Pattern Recognition},
  pages     = {8317--8326},
  year      = {2019}
}

@inproceedings{biten2019stvqa,
  title     = {Scene Text Visual Question Answering},
  author    = {Biten, Ali Furkan and Tito, Ruben and Mafla, Andres and Gomez, Lluis and Rusi{\~n}ol, Mar{\c{c}}al and Valveny, Ernest and Jawahar, C. V. and Karatzas, Dimosthenis},
  booktitle = {Proceedings of the IEEE/CVF International Conference on Computer Vision},
  pages     = {4291--4301},
  year      = {2019}
}

@article{liu2024ocrbench,
  title   = {{OCRBench}: On the Hidden Mystery of {OCR} in Large Multimodal Models},
  author  = {Liu, Yuliang and Li, Zhang and Huang, Mingxin and Yang, Biao and Yu, Wenwen and Li, Chunyuan and Yin, Xu-Cheng and Liu, Cheng-Lin and Jin, Lianwen and Bai, Xiang},
  journal = {Science China Information Sciences},
  volume  = {67},
  number  = {12},
  pages   = {220102},
  year    = {2024},
  doi     = {10.1007/s11432-024-4235-6}
}

@misc{bai2025qwen25vl,
  title         = {{Qwen2.5-VL} Technical Report},
  author        = {Bai, Shuai and Chen, Keqin and Liu, Xuejing and Wang, Jialin and Ge, Wenbin and Song, Sibo and Dang, Kai and Wang, Peng and Wang, Shijie and Tang, Jun and Zhong, Humen and Zhu, Yuanzhi and Yang, Mingkun and Li, Zhaohai and Wan, Jianqiang and Wang, Pengfei and Ding, Wei and Fu, Zheren and Xu, Yiheng and Ye, Jiabo and Zhang, Xi and Xie, Tianbao and Cheng, Zesen and Zhang, Hang and Yang, Zhibo and Xu, Haiyang and Lin, Junyang},
  year          = {2025},
  eprint        = {2502.13923},
  archivePrefix = {arXiv},
  primaryClass  = {cs.CV},
  doi           = {10.48550/arXiv.2502.13923}
}

@article{li2025llavaonevision,
  title   = {{LLaVA-OneVision}: Easy Visual Task Transfer},
  author  = {Li, Bo and Zhang, Yuanhan and Guo, Dong and Zhang, Renrui and Li, Feng and Zhang, Hao and Zhang, Kaichen and Zhang, Peiyuan and Li, Yanwei and Liu, Ziwei and Li, Chunyuan},
  journal = {Transactions on Machine Learning Research},
  year    = {2025}
}

@misc{wang2025internvl35,
  title         = {{InternVL3.5}: Advancing Open-Source Multimodal Models in Versatility, Reasoning, and Efficiency},
  author        = {Wang, Weiyun and Gao, Zhangwei and Gu, Lixin and Pu, Hengjun and Cui, Long and Wei, Xingguang and Liu, Zhaoyang and Jing, Linglin and Ye, Shenglong and Shao, Jie and Wang, Zhaokai and Chen, Zhe and Zhang, Hongjie and Yang, Ganlin and Wang, Haomin and Wei, Qi and Yin, Jinhui and Li, Wenhao and Cui, Erfei and Chen, Guanzhou and Ding, Zichen and Tian, Changyao and Wu, Zhenyu and Xie, Jingjing and Li, Zehao and Yang, Bowen and Duan, Yuchen and Wang, Xuehui and Hou, Zhi and Hao, Haoran and Zhang, Tianyi and Li, Songze and Zhao, Xiangyu and Duan, Haodong and Deng, Nianchen and Fu, Bin and He, Yinan and Wang, Yi and He, Conghui and Shi, Botian and He, Junjun and Xiong, Yingtong and Lv, Han and Wu, Lijun and Shao, Wenqi and Zhang, Kaipeng and Deng, Huipeng and Qi, Biqing and Ge, Jiaye and Guo, Qipeng and Zhang, Wenwei and Zhang, Songyang and Cao, Maosong and Lin, Junyao and Tang, Kexian and Gao, Jianfei and Huang, Haian and Gu, Yuzhe and Lyu, Chengqi and Tang, Huanze and Wang, Rui and Lv, Haijun and Ouyang, Wanli and Wang, Limin and Dou, Min and Zhu, Xizhou and Lu, Tong and Lin, Dahua and Dai, Jifeng and Su, Weijie and Zhou, Bowen and Chen, Kai and Qiao, Yu and Wang, Wenhai and Luo, Gen},
  year          = {2025},
  eprint        = {2508.18265},
  archivePrefix = {arXiv},
  primaryClass  = {cs.CV},
  doi           = {10.48550/arXiv.2508.18265}
}

@inproceedings{azuma2023defenseprefix,
  title     = {Defense-Prefix for Preventing Typographic Attacks on {CLIP}},
  author    = {Azuma, Hiroki and Matsui, Yusuke},
  booktitle = {Proceedings of the IEEE/CVF International Conference on Computer Vision Workshops},
  pages     = {3644--3653},
  year      = {2023}
}

@inproceedings{ilharco2022paint,
  title     = {Patching Open-Vocabulary Models by Interpolating Weights},
  author    = {Ilharco, Gabriel and Wortsman, Mitchell and Gadre, Samir Yitzhak and Song, Shuran and Hajishirzi, Hannaneh and Kornblith, Simon and Farhadi, Ali and Schmidt, Ludwig},
  booktitle = {Advances in Neural Information Processing Systems},
  volume    = {35},
  pages     = {29262--29277},
  year      = {2022}
}

@inproceedings{hufe2026dyslexify,
  title     = {{Dyslexify}: A Mechanistic Defense Against Typographic Attacks in {CLIP}},
  author    = {Hufe, Lorenz and Venhoff, Constantin and Purelku, Erblina and Dreyer, Maximilian and Lapuschkin, Sebastian and Samek, Wojciech},
  booktitle = {International Conference on Learning Representations},
  year      = {2026},
  url       = {https://openreview.net/forum?id=UI7mbsIZeN}
}

@inproceedings{qraitem2025webartifact,
  title     = {Web Artifact Attacks Disrupt Vision Language Models},
  author    = {Qraitem, Maan and Teterwak, Piotr and Saenko, Kate and Plummer, Bryan A.},
  booktitle = {Proceedings of the IEEE/CVF International Conference on Computer Vision},
  pages     = {1048--1057},
  year      = {2025}
}

@misc{qraitem2024selfgenerated,
  title         = {Vision-{LLM}s Can Fool Themselves with Self-Generated Typographic Attacks},
  author        = {Qraitem, Maan and Tasnim, Nazia and Teterwak, Piotr and Saenko, Kate and Plummer, Bryan A.},
  year          = {2024},
  eprint        = {2402.00626},
  archivePrefix = {arXiv},
  primaryClass  = {cs.CV},
  doi           = {10.48550/arXiv.2402.00626}
}

@article{goh2021multimodal,
  title   = {Multimodal Neurons in Artificial Neural Networks},
  author  = {Goh, Gabriel and Cammarata, Nick and Voss, Chelsea and Carter, Shan and Petrov, Michael and Schubert, Ludwig and Radford, Alec and Olah, Chris},
  journal = {Distill},
  year    = {2021},
  doi     = {10.23915/distill.00030},
  url     = {https://distill.pub/2021/multimodal-neurons/}
}

@inproceedings{materzynska2022disentangling,
  title     = {Disentangling Visual and Written Concepts in {CLIP}},
  author    = {Materzy{\'n}ska, Joanna and Torralba, Antonio and Bau, David},
  booktitle = {Proceedings of the IEEE/CVF Conference on Computer Vision and Pattern Recognition},
  pages     = {16410--16419},
  year      = {2022}
}

@inproceedings{wang2025multiimage,
  title     = {Typographic Attacks in a Multi-Image Setting},
  author    = {Wang, Xiaomeng and Zhao, Zhengyu and Larson, Martha},
  booktitle = {Proceedings of the 2025 Conference of the Nations of the Americas Chapter of the Association for Computational Linguistics: Human Language Technologies},
  pages     = {12594--12604},
  publisher = {Association for Computational Linguistics},
  year      = {2025},
  doi       = {10.18653/v1/2025.naacl-long.626},
  url       = {https://aclanthology.org/2025.naacl-long.626/}
}

@inproceedings{gong2025figstep,
  title     = {{FigStep}: Jailbreaking Large Vision-Language Models via Typographic Visual Prompts},
  author    = {Gong, Yichen and Ran, Delong and Liu, Jinyuan and Wang, Conglei and Cong, Tianshuo and Wang, Anyu and Duan, Sisi and Wang, Xiaoyun},
  booktitle = {Proceedings of the AAAI Conference on Artificial Intelligence},
  pages     = {23951--23959},
  year      = {2025},
  doi       = {10.1609/aaai.v39i22.34568}
}

@misc{kimura2024goalhijacking,
  title         = {Empirical Analysis of Large Vision-Language Models against Goal Hijacking via Visual Prompt Injection},
  author        = {Kimura, Subaru and Tanaka, Ryota and Miyawaki, Shumpei and Suzuki, Jun and Sakaguchi, Keisuke},
  year          = {2024},
  eprint        = {2408.03554},
  archivePrefix = {arXiv},
  primaryClass  = {cs.CL},
  doi           = {10.48550/arXiv.2408.03554}
}

@misc{openai2025gpt41,
  title        = {Introducing {GPT-4.1} in the {API}},
  author       = {{OpenAI}},
  year         = {2025},
  howpublished = {\url{https://openai.com/index/gpt-4-1/}},
  note         = {Accessed: 2026-07-24}
}

@inproceedings{mathew2021docvqa,
  title     = {{DocVQA}: A Dataset for {VQA} on Document Images},
  author    = {Mathew, Minesh and Karatzas, Dimosthenis and Jawahar, C. V.},
  booktitle = {Proceedings of the IEEE/CVF Winter Conference on Applications of Computer Vision},
  pages     = {2200--2209},
  year      = {2021}
}

@inproceedings{masry2022chartqa,
  title     = {{ChartQA}: A Benchmark for Question Answering about Charts with Visual and Logical Reasoning},
  author    = {Masry, Ahmed and Long, Do Xuan and Tan, Jia Qing and Joty, Shafiq and Hoque, Enamul},
  booktitle = {Findings of the Association for Computational Linguistics: ACL 2022},
  pages     = {2263--2279},
  publisher = {Association for Computational Linguistics},
  year      = {2022},
  doi       = {10.18653/v1/2022.findings-acl.177},
  url       = {https://aclanthology.org/2022.findings-acl.177/}
}

@inproceedings{chen2024internvl,
  title     = {{InternVL}: Scaling up Vision Foundation Models and Aligning for Generic Visual-Linguistic Tasks},
  author    = {Chen, Zhe and Wu, Jiannan and Wang, Wenhai and Su, Weijie and Chen, Guo and Xing, Sen and Zhong, Muyan and Zhang, Qinglong and Zhu, Xizhou and Lu, Lewei and Li, Bin and Luo, Ping and Lu, Tong and Qiao, Yu and Dai, Jifeng},
  booktitle = {Proceedings of the IEEE/CVF Conference on Computer Vision and Pattern Recognition},
  pages     = {24185--24198},
  year      = {2024}
}

@inproceedings{suvorov2022lama,
  title     = {Resolution-Robust Large Mask Inpainting with Fourier Convolutions},
  author    = {Suvorov, Roman and Logacheva, Elizaveta and Mashikhin, Anton and Remizova, Anastasia and Ashukha, Arsenii and Silvestrov, Aleksei and Kong, Naejin and Goka, Harshith and Park, Kiwoong and Lempitsky, Victor},
  booktitle = {Proceedings of the IEEE/CVF Winter Conference on Applications of Computer Vision},
  pages     = {2149--2159},
  year      = {2022}
}

@inproceedings{fang2025rsste,
  title     = {Recognition-Synergistic Scene Text Editing},
  author    = {Fang, Zhengyao and Lyu, Pengyuan and Wu, Jingjing and Zhang, Chengquan and Yu, Jun and Lu, Guangming and Pei, Wenjie},
  booktitle = {Proceedings of the IEEE/CVF Conference on Computer Vision and Pattern Recognition},
  pages     = {13104--13113},
  year      = {2025}
}

@misc{zhang2026ppocrv6,
  title         = {{PP-OCRv6}: From 1.5M to 34.5M Parameters, Surpassing Billion-Scale {VLMs} on {OCR} Tasks},
  author        = {Zhang, Yubo and Wang, Xueqing and Lin, Manhui and Zhang, Yue and Deng, Penglongyi and Sun, Ting and Gao, Tingquan and Zhang, Zelun and Liu, Jiaxuan and Zhou, Changda and Liu, Hongen and Liang, Suyin and Cui, Cheng and Liu, Yi and Yu, Dianhai and Ma, Yanjun},
  year          = {2026},
  eprint        = {2606.13108},
  archivePrefix = {arXiv},
  primaryClass  = {cs.CV},
  doi           = {10.48550/arXiv.2606.13108}
}

@techreport{maji2013aircraft,
  title       = {Fine-Grained Visual Classification of Aircraft},
  author      = {Maji, Subhransu and Kannala, Juho and Rahtu, Esa and Blaschko, Matthew and Vedaldi, Andrea},
  institution = {arXiv},
  year        = {2013},
  number      = {arXiv:1306.5151}
}

@inproceedings{bossard2014food101,
  title     = {{Food-101}: Mining Discriminative Components with Random Forests},
  author    = {Bossard, Lukas and Guillaumin, Matthieu and Van Gool, Luc},
  booktitle = {European Conference on Computer Vision},
  pages     = {446--461},
  year      = {2014}
}

@misc{shekhar2021imagenet100,
  author       = {Shekhar, Ambesh},
  title        = {ImageNet100: A Sample of ImageNet Classes},
  year         = {2021},
  howpublished = {Kaggle Dataset, \url{https://www.kaggle.com/datasets/ambityga/imagenet100}}
}

\appendix

\section{Additional Details of the Motivation Studies}
\label{app:motivation_protocols}

\subsection{Appearance and Semantic Modifications}
The 600-sample study uses 200 attacked pairs from each of SCAM, SceneTAP, and SELF. SCAM contains equal synthetic and real subsets, SceneTAP is stratified by question type, and SELF is balanced across its five source categories. Pilot samples are excluded. Every operator is restricted to the same inserted-text region, obtained by aligning the clean--attack difference mask with frozen OCR boxes. Qwen2.5-VL-7B-Instruct and LLaVA-OneVision-8B answer the same fixed multiple-choice question under every condition. Table~\ref{tab:supp_controlled_modifications} gives the fixed operators used in Figure~2 of the main paper.

\begin{table}[ht]
\centering
\small
\setlength{\tabcolsep}{3pt}
\begin{tabular}{@{}lp{0.72\columnwidth}@{}}
\toprule
\multicolumn{2}{@{}l}{\emph{Appearance modifications}} \\
\midrule
Blur & Gaussian blur blended with $\alpha=0.65$; kernel $\approx0.55$ of median text height \\
Low contrast & retain 0.65 of each text-stroke value and mix 0.35 local background \\
Font change & erase and redraw the same string in DejaVuSerif-Italic \\
Rotation & affine transform with $6^\circ$ rotation, 0.04 shear, and 0.98 scale \\
Downscale & scale the original text layer by 0.5 and center it in the source box \\
\addlinespace[2pt]
\multicolumn{2}{@{}l}{\emph{Semantic modifications}} \\
\midrule
Neutral word & replace with a task-level abstract term, e.g., \emph{object}, \emph{color}, or \emph{number} \\
Pseudo-text & substitute letters from \texttt{qzxvkjw} and digits randomly while preserving length and case \\
Char. corruption & replace 40\% of alphanumeric positions with symbols \\
Delete & inpaint the selected support mask with LaMa \\
Blank patch & fill each OCR rectangle with its surrounding-ring mean color \\
\bottomrule
\end{tabular}
\caption{Controlled modification settings used for Observation~1.}
\label{tab:supp_controlled_modifications}
\end{table}
\FloatBarrier

\subsection{Construction of the Four Semantic Conditions}
Table~\ref{tab:supp_qi_construction} summarizes how the four semantic variants defined by query relevance (Q) and image relevance (I) are constructed and accepted.

\begin{table}[ht]
\centering
\small
\setlength{\tabcolsep}{3pt}
\begin{tabular}{@{}lp{0.75\columnwidth}@{}}
\toprule
Condition & Construction and acceptance rule \\
\midrule
Q1-I1 & Generate non-copy proposals in the queried answer category from the text-removed image, question, task type, and ground-truth answer; select one supported by the queried image content. \\
Q1-I0 & Form a fixed task-category pool, remove ground-truth-answer and attack-text surfaces, and select a same-category concept that is false or unsupported for the queried image content. \\
Q0-I1 & Generate 8--12 visible concepts from seven visual dimensions; remove overlaps with the query target, ground-truth answer, attack text, and queried dimension, then select a visible query-irrelevant concept. \\
Q0-I0 & Select deterministically from a frozen abstract pool after removing overlaps with the question, ground-truth answer, attack text, and the other Q/I conditions. \\
\bottomrule
\end{tabular}
\caption{Construction of the four semantic conditions in Observation~2.}
\label{tab:supp_qi_construction}
\end{table}
\FloatBarrier

The four variants are generated from 2,237 eligible non-complex SCAM and SceneTAP pairs and evaluated on the common 1,810-pair subset for which all conditions are available. Answer options are removed, and all conditions use the same concise-answer prompt. Qwen2.5-VL-7B-Instruct performs the image-conditioned proposal and selection steps, while Q0-I0 is selected deterministically. Rule-based checks reject copied anchors, invalid categories, cross-condition duplicates, and malformed surfaces. The protocol is frozen after validation on two disjoint 175-sample sets and a sampled human audit, and the resulting edits are held fixed across target models. Responses unresolved by deterministic normalization are judged without access to the semantic condition.

\section{Additional Implementation Details of QuISE}
\label{app:quise_implementation}

\subsection{Defense-Time Localization}
At inference time, the target VLM first recognizes candidate text in the input image, while PaddleOCR provides spatial text boxes. Normalized text matching associates the VLM-recognized strings with individual or merged adjacent OCR boxes. Given the image, query, answer options, and grounded candidates, the same target VLM selects all text regions that may influence its answer. Only these grounded regions are passed to semantic editing. If no influential text can be reliably localized, or any localization stage fails, QuISE retains the input-image answer.

\subsection{Defense-Time Q0-I0 Selection}
For each localized text group, QuISE constructs a deterministic, geometry-compatible shortlist from a frozen abstract candidate pool after removing candidates that overlap with the source text, query, or answer options. The target VLM then evaluates every candidate against the query and the text-removed image, accepting it as Q0-I0 only when it is judged irrelevant to both; uncertain candidates are rejected. If fewer than two candidates are accepted, QuISE performs one additional attempt with a disjoint shortlist. It selects two verified candidates, preferring different semantic classes, and uses them to create the two edited images. The ground-truth answer and the input-image answer are never provided during selection. Insufficient verified candidates or any subsequent editing failure triggers the original-answer fallback.

\end{document}